\documentclass[conference]{IEEEtran}
\IEEEoverridecommandlockouts
\usepackage{cite}
\usepackage{amsmath,amssymb,amsfonts}
\usepackage{algorithmic}
\usepackage{graphicx}
\usepackage{textcomp}
\usepackage[table]{xcolor}
\usepackage{multirow}
\usepackage{booktabs}
\usepackage{siunitx}
\usepackage{nicematrix}

\usepackage{url}

\DeclareUnicodeCharacter{2009}{\,} 
\def\BibTeX{{\rm B\kern-.05em{\sc i\kern-.025em b}\kern-.08em
    T\kern-.1667em\lower.7ex\hbox{E}\kern-.125emX}}

\begin{document}
\title{Experiential-Informed Data Reconstruction for Fishery Sustainability and Policies in the Azores
}


\author{\IEEEauthorblockN{Brenda Nogueira}
\IEEEauthorblockA{\textit{Faculty of Science} \\
\textit{University of Porto}\\
Porto, Portugal \\
0000-0003-1936-9863
}
\and
\IEEEauthorblockN{Gui M. Menezes}
\IEEEauthorblockA{\textit{Institute of Marine Sciences-OKEANOS} \\
\textit{University of the Azores}\\
Horta, Azores, Portugal \\
gui.mm.menezes@uac.pt}
\and
\IEEEauthorblockN{Nuno Moniz}
\IEEEauthorblockA{\textit{Lucy Family Institute for Data \& Society} \\
\textit{University of Notre Dame}\\
Indiana, USA \\
nuno.moniz@nd.edu}
\and
\IEEEauthorblockN{Rita P. Ribeiro}
\IEEEauthorblockA{\textit{Faculty of Science} \\
\textit{University of Porto}\\
Porto, Portugal \\
rpribeiro@fc.up.pt}

}

\maketitle

\begin{abstract}
Fishery analysis is critical in maintaining the long-term sustainability of species and the livelihoods of millions of people who depend on fishing for food and income. The fishing gear, or metier, is a key factor significantly impacting marine habitats, selectively targeting species and fish sizes. Analysis of commercial catches or landings by metier in fishery stock assessment and management is crucial, providing robust estimates of fishing efforts and their impact on marine ecosystems. In this paper, we focus on a unique data set from the Azores' fishing data collection programs between 2010 and 2017, where little information on metiers is available and sparse throughout our timeline. Our main objective is to tackle the task of data set reconstruction, leveraging domain knowledge and machine learning methods to retrieve or associate metier-related information to each fish landing. We empirically validate the feasibility of this task using a diverse set of modeling approaches and demonstrate how it provides new insights into different fisheries' behavior and the impact of metiers over time, which are essential for future fish population assessments, management, and conservation efforts.
\end{abstract}

\begin{IEEEkeywords}
sustainability, fishery data, data set reconstruction, machine learning, evidence-based policy
\end{IEEEkeywords}

\section{Introduction}\label{sec:intro}

The global demand for seafood has increased substantially, with an estimated 179 million tons of fish production worldwide in 2018, of which 156 million tons were used for human consumption -- an annual supply of 20.5 kg per capita~\cite{i1}. However, this high demand has also had an unprecedented impact on aquatic ecosystems and is believed to have caused a reduction in ocean biomass content of up to 80\%~\cite{i2}. The proportion of fish stocks within biologically sustainable levels decreased from 90\% in 1974 to 65.8\% in 2017~\cite{i1}. Furthermore, recent assessments have presented worrisome statistics regarding fish and shellfish stocks. Out of the 397 commercially exploited stocks, a staggering 69\% were classified as overfished, reflecting the consequences of excessive fishing activities. Additionally, 51\% of the stocks were found to be operating beyond safe biological limits, putting their long-term sustainability at risk. These findings become even more concerning when considering that only 12\% of the stocks met the established guidelines outlined by the Common Fisheries Policy (CFP), which governs fisheries in the European Union seas~\cite{i3}. Such data emphasizes the pressing need for immediate measures to address overfishing and ensure the implementation of sustainable practices for the preservation and future of these vital marine resources.

Portugal has Europe's 5th largest exclusive economic zone (EEZ), the 3rd largest of the EU, and the 20th largest EEZ globally, at 1,727,408 km2. The Autonomous Region of Azores (Portugal), a group of 9 islands spread over 600 km wide, is a major contributor to the size of the country's EEZ. In the Azores, fisheries are considered artisanal and small-scale in nature, with a multisegmented fleet, targeting multiple species with a wide range of fishing gears or metiers~\cite{i4}, managed under the CFP and by national and regional management policies~\cite{i5}. Its annual landings in the last ten years are about 11.000 tons and valued at about 33M€. 

The Region implemented many technical and spatial measures, monitoring scientific surveys and the fishing effort, biological data, or the size composition of the species landings, are regularly collected under the national DCF (the PNRD). Despite efforts, information is usually deficient or inadequate for most species. Among the 138 species that landed in the region between 2009-2019, twenty-two (18 fishes, two molluscs, and two crustaceans) were selected as priority stocks according to the FAO\footnote{FAO -- Food and Agriculture Organization of the United Nations} and ICES\footnote{ICES -- International Council for the Exploration of the Sea} criteria~\cite{i8}. Most of these showed a decreasing trend in their abundances, and only four stocks are currently assessed using data-limited approaches~\cite{i8}. However, The uncertainty inherent to data-limited stock assessments, such as the existing ones, can compromise the ability to inform management~\cite{i9,i10,i11}. In addition, the more robust population assessment methods require considerable data, which are virtually impossible to obtain for all species, and limited fisheries data~\cite{i12}, such as our study case.

In this paper, we leverage domain knowledge and machine learning tools to improve the quality of the fisheries database in the Azores, fill the information gaps, improve the fish stock assessment modeling, and decrease its uncertainty, with the ultimate goal of improving advice and sustainable management of the wild fish populations. Specifically, we explore combinations of data pre-processing strategies and feature engineering methods to cope with two main challenges: multi-class imbalance and lack of contextual information in the original data.  

The remainder of this paper is organized as follows. Section ~\ref{sec:background} discusses related works, accompanied by motivating examples. In Section ~\ref{sec:data}, the data is introduced, along with a description of the preparation and preprocessing steps undertaken, as well as a concise analysis that serves as a motivation for the study. Section ~\ref{sec:expanalysis} outlines the evaluation metrics employed for the time series data set and presents the machine learning models and strategies utilized, along with the corresponding results obtained. Section \ref{sec:discussion} consists of a comprehensive discussion of the results, followed by the conclusions presented in Section \ref{sec:conclusion}. 

\section{Background}\label{sec:background}

Incomplete datasets, where some samples lack attribute values, have long been a challenge in research. Traditional imputation techniques like mean imputation or last observation carried forward have limitations as they oversimplify the data, leading to biased estimates and distorted conclusions. Researchers have turned to machine learning algorithms for missing data reconstruction to address this.

The utility of machine learning in filling in missing information has been demonstrated in various domains. For instance, in a study focused on gene network reconstruction \cite{b1}, microarray experiments were used to identify gene interrelations. A supervised learning approach utilizing decision-tree-related classifiers was employed to predict gene expression based on other gene expression data. In another study ~\cite{b2}, the authors focus on compensating for missing data in electric power datasets for improved energy management systems. It compares the performance of statistical methods (ARIMA and LI) and machine learning methods (K-NN, MLP, and SVR) for data imputation. Using a two-year dataset from Taiwan, the researchers find that machine learning methods generally outperform statistical methods. 

Machine learning offers advantages, particularly in temporal data. Traditional methods struggle with the dynamic patterns and dependencies found in temporal data. In~\cite{b3}, the Engdahl-van der Hilst-Buland (EHB) Bulletin of Hypocentres and associated travel time residuals were reconstructed using updated procedures. This resulted in improved maps of seismicity, benefiting global seismicity studies and tomographic inversions.

Marine ecosystem research and conservation present unique challenges due to their complex nature and data collection across various scales \cite{b4}. Machine learning automates routine tasks in marine science, enabling data-driven decision-making and uncovering hidden patterns in large datasets \cite{b5}.

Our work holds particular significance in the fisheries domain as it showcases the effectiveness of utilizing machine learning to address the issue of missing data in fishery landings on a temporal relevant scale. This research serves as a valuable contribution to the field, highlighting the efficacy of machine learning in overcoming missing data challenges in a dataset of critical importance to fisheries.

\section{Fisheries Data (2010-2017)}\label{sec:data}



This section provides an overview of the data utilized in our study. Specifically, we employed the LOTAÇOR/OKEANOS-UAc daily landings dataset and the PNRD/OKEANOS-UAc inquiries database. Samplers randomly collect these inquiries during the process of fishery landings in the main fishing harbors in the Azores, including São Mateus, Praia da Vitória, Rabo de Peixe, Santa Cruz (Faial), Ponta Delgada (São Miguel), Povoação, Madalena, Vila do Porto, and Angra do Heroismo. These inquiries are used to get information from fisheries activities, mainly about the fishing gears used, the fishing effort, the baits, the characteristics of the fishing gears used, or the metiers. 

The data used in this paper covers the period between 2010 to 2017 and comprises 13 categories of fishing gears -- referred to as \textbf{metiers}, detailed in Table~\ref{tab_metiers}.

\begin{table*}[!ht]
\caption{Metier's description and number of landings}
\begin{center}
\begin{tabular}{|c|c|c|c|} 
 \hline
 Metier & Description & Quantity & Percentage (\%)\\ 
 \hline
 FPO-CRU & Crustacean traps & 2 & 0.0143 \\
 \hline
 FPO-PB & Fish traps & 46 & 0.330 \\
 \hline
 GNS-PB & Coastal gillnets & 365 & 2.62 \\
 \hline
 LHP-CEF & Handline jigging for catching squids & 3316 & 23.8 \\
 \hline
 LHP-PB & Bottom fish handlines & 4532 & 32.5 \\
 \hline
 LHP-PBC & Pole-and-line and coastal trolling for small pelagics & 22 & 0.158 \\
 \hline
 LHP-TUN & Pole-and-line for tuna species & 1019 & 7.30 \\
 \hline
 LLD-GPP & Surface drifting longline for large migratory pelagics & 31 & 0.222 \\
 \hline
 LLD-PP & Deep-water drift bottom longline & 109 & 0.781 \\
 \hline
LLS-DEEP & Deep-water non-drifting bottom longline & 231 & 1.66 \\
\hline
LLS-PD & Bottom longline & 2732 & 19.6 \\
\hline
PS-PB & Lifting nets for small coastal fishes & 43 & 0.308 \\
\hline
PS-PPP & Purse seine nets for small pelagic fish & 1507 & 10.8 \\
\hline
\end{tabular}
\end{center}
\label{tab_metiers}
\end{table*}

The data cleaning process involved several steps, divided into two stages: preparation and pre-processing. The first stage aims at ensuring our data's accuracy and integrity. The second stage ensures that each vessel's daily fish landings are considered an independent sample, essential for the ensuing model development processes. They are detailed as follows.

\begin{itemize}
    \item \textbf{Preparation}: The resulting data set contains 11 features (described in Table \ref{tab_variables}) and 33,895 samples.
    \begin{itemize}
        \item Feature selection to remove variables with a high percentage of missing data, ensuring that our data was complete and reliable;
        \item Data set analysis for sensitive information -- features containing information that could be used for re-identification; these were also removed;
        \item Data transformation step to replace the name of metiers' classes, LHM-CEF and LHM-PB, with LHP-CEF and LHP-PB, respectively, due to a change in designation at the beginning of 2017;
        \item Concerning the boat registration code feature, we only retained each boat's "C" or "L" classification code. "C" stands for fishing vessels with cabins, and "L" for smaller, open-deck fishing vessels.
    \end{itemize}
    \item \textbf{Pre-Processing}: The final data set contains 13,955 landings (cases) and 138 features.
    \begin{itemize}
        \item Information concerning landings was consolidated to create one row per landing, i.e., from a row per landing/species to a row per landing;
        \item One-hot encoding applied to each categorical feature, except for the species classification feature.
    \end{itemize}
\end{itemize}

\begin{table}
\caption{Features description}
\begin{center}
\begin{tabular}{|c|c|} 
 \hline
 \textbf{Feature} & \textbf{Description} \\ 
 \hline
 \text{Id\_landing} & \text{daily vessel landing Id} \\ 
 \hline
 Data\_landing & Date of the landing \\
 \hline 
 Ilha & The island where the landing took place \\
 \hline
 Porto & Fish port of the landing \\
 \hline
 Vulgar & Common fish name \\
 \hline
 Classificação & Major fish species group\\
 \hline
 Peso & Weight of the landing per species  \\
 \hline
 Matricula & Boat registration code \\
 \hline
 Classe\_com & Boat size group code\\
 \hline
 Compff & Boat length \\
 \hline
 Metier & Type of fishing gear \\
 \hline
\end{tabular}
\end{center}
\label{tab_variables}
\end{table}

\subsection{Data Analysis}\label{subsec:analysis}
 
To conduct a comprehensive analysis of fish population dynamics and sustainability of each species, it is crucial to know the biology and ecology of the species (e.g., growth, mortality, reproduction features, etc.), to study the fishing operation regimes and the metiers involved, they collect data on fishing catches (or landings) per species, the fishing geographic distribution, the fishing effort and how these patterns evolve. This information provides insights into the dynamics and exploitation rates of each species or species group over time.

\begin{figure*}[htbp]
\centerline{\includegraphics[width=1.0\linewidth]{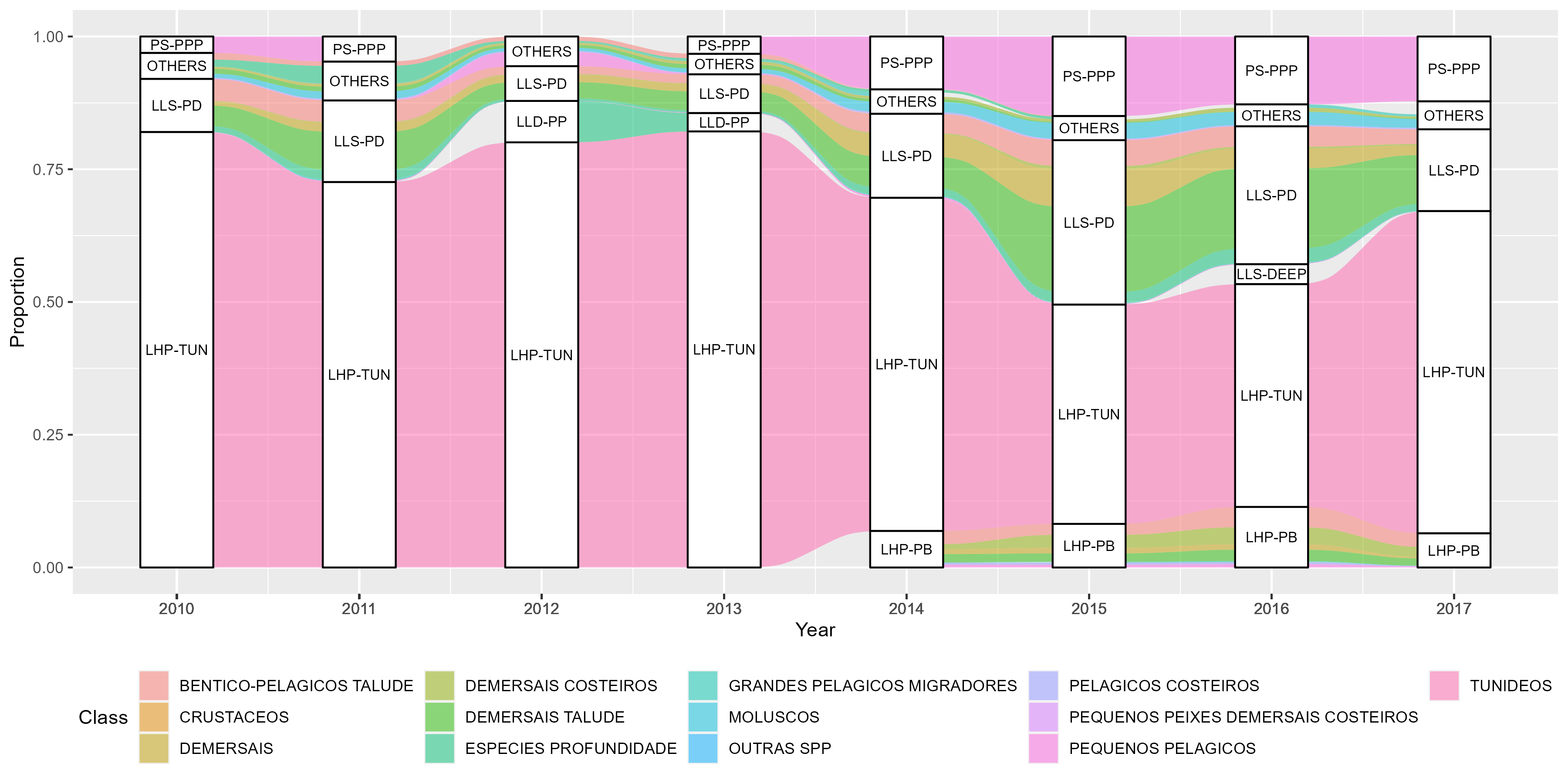}}
\caption{Distribution of major fish groups over the years per metier.}
\label{fig_dist_over_years}
\end{figure*}

One important question in the fisheries industry study is the species caught across different metiers over time. Figure \ref{fig_dist_over_years} shows the proportion of major fish groups weight caught for each metier over the years in the Azores, and a trend has emerged where significant species such as tunas (\textit{tunideos}) have declined in relevance, allowing other species to become more prominent. Tunas are migratory fish species, and catches may strongly fluctuate from one year to another, and the occurrence of different tuna species also varies mainly due to environmental causes.

\begin{figure*}[htbp]
\centerline{\includegraphics[width=1.0\linewidth]{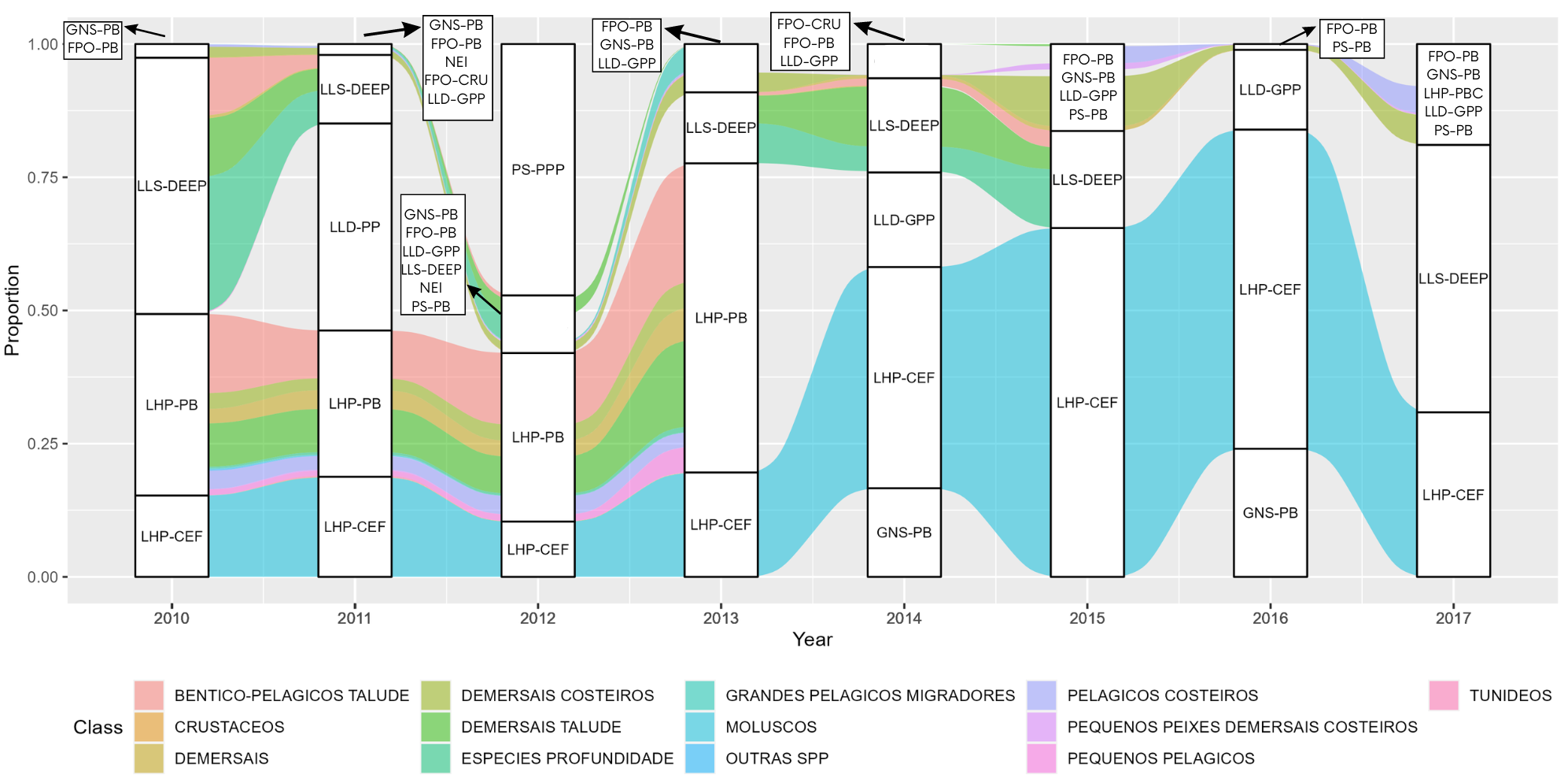}}
\caption{Distribution of major fish groups over the years per metier: small fish classes}
\label{fig_small_over_years}
\end{figure*}

The image in Figure~\ref{fig_dist_over_years} shows that the "Others" section represents the metiers that have a relatively small proportion in the overall fishery. To better understand the impact of these metiers on the less significant fish species, we have zoomed in on this section and presented the data in Figure~\ref{fig_small_over_years}. By analyzing the figure depicting the lesser-known metiers, we can observe a notable shift in representation. Some of these metiers become more prominent in the other image, disappearing from the current one. This observation suggests a potential improvement in the distribution of fishery activities among species in the latter image, indicating a more balanced and diversified utilization of fishing techniques

\begin{figure*}[htbp]
\centerline{\includegraphics[width=1.0\linewidth,height=0.25\textheight]{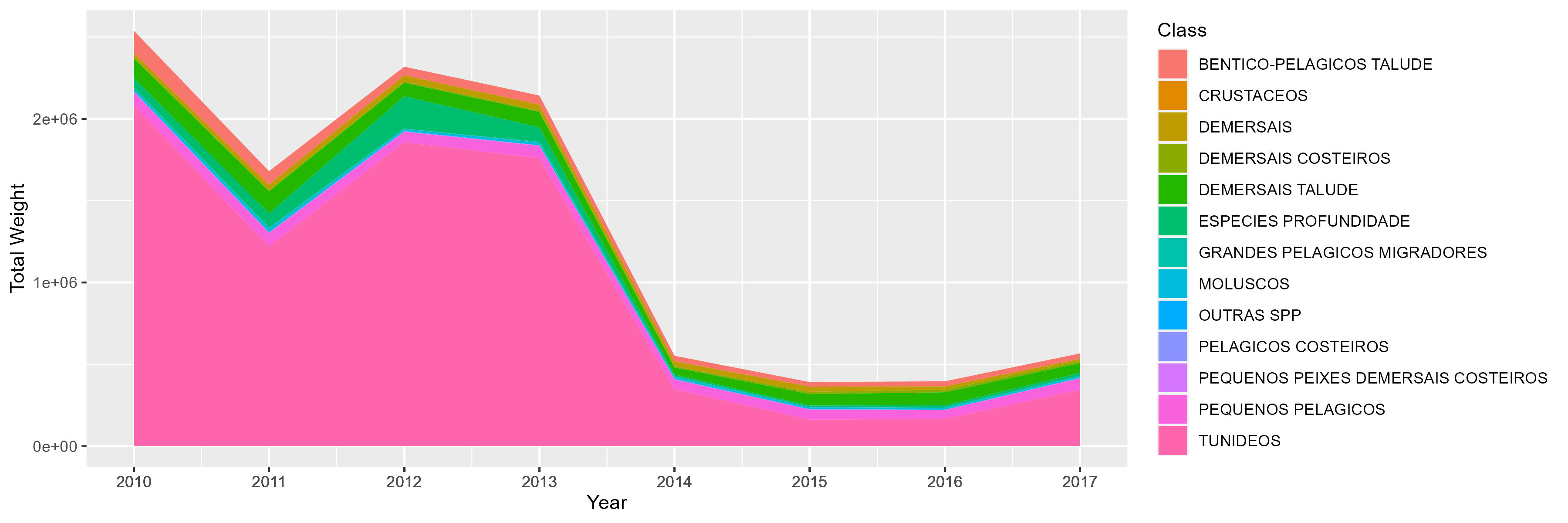}}
\caption{Total weight of each fish classification over the years}
\label{fig_area_over_years}
\end{figure*}

To gain a comprehensive understanding of each species' dynamics and the fishery industry's overall sustainability, it is also crucial to know the species' total catches (total weight) and how it changes over time. This information is presented in Figure~\ref{fig_area_over_years}. It is evident in this case that the total weight of fish caught has decreased over the years, indicative of a potential intensive fishing or overfishing issue. The variations in catches are caused by various management practices, such as implementing Total Allowable Annual Catches (TACs), reducing fishing efforts, or changing the number of fishing vessels, among other factors. Therefore, it would be interesting to calculate catch-per-unit-effort (CPUE) using the effort hours. CPUE provides more reliable indicators of population abundance and is commonly employed in fish stock assessment models.   

However, obtaining robust data on species CPUE of commercial or recreational fisheries is difficult and error-prone. 
Information is mainly obtained from inquiries sampling programs and raised to totals following statistical procedures assuming the accomplishment of sampling statistical assumptions in terms of representativeness of the samples of the variables of interest (e.g., representativeness of species landings, fishing vessels, fishing metiers, fish ports, etc.). For example, associating a specific fishing gear to a specific fish landing to obtain a good estimation of the fishing effort (e.g., number of hooks, hours fishing, trawl area, etc.) is quite limited. Fishing survey programs have significant gaps in information and coverage, particularly regarding the metiers or fishing efforts used in fishing. Frequently, this lack of detailed data can hinder more comprehensive studies and robust fisheries assessments, leading to uncertain results.

Our work addresses this issue by reconstructing the missing data in existing Azores fisheries surveys. The reconstruction of the missing data of the LOTAÇOR/OKEANOS-UAc fish landings database will fill the gaps and provide more accurate information for future stock assessment modeling processes resulting in better and less uncertain advice for fishing management.

\section{Experimental Analysis}\label{sec:expanalysis}

Accurately predicting missing information in data analysis is a challenging and complex task with various applications in various fields. We explored six questions regarding the optimization and improvement of machine learning models on a dataset: 

\begin{enumerate}
    \item How does the performance of different machine learning methods vary on our dataset, and how can we identify the ideal time frame for making accurate predictions on our time series data? What impact does this have on the precision and stability of our models?
    \item To what extent can contextual variables enhance the accuracy and robustness of our predictions, and how do we select the most relevant features to include?
    \item What insights can be gained by selectively removing features from our dataset, and how can we leverage this information to improve our models?
    \item How does rebalancing an imbalanced dataset affect the quality and representativeness of our training data, and what are the trade-offs involved in different rebalancing approaches?
    \item How can ensemble learning techniques be used to enhance the accuracy and robustness of our models, and what are the best practices for incorporating these methods into our workflow?
    \item What are the alternative metrics, apart from accuracy, can effectively reflect the performance of an imbalanced classification task?
\end{enumerate}

First, we explore the effectiveness of several supervised machine learning methods in predicting the metiers of fish landings. We evaluate different train and test evaluation periods to determine the optimal timeframe for prediction. Our analysis aims to identify the optimal balance between the size of the training set and the prediction horizon to ensure that our model is accurate and timely.

In addition, we examine the impact of adding context-related variables on the machine learning model's performance. We hypothesize that incorporating such variables can improve our model's accuracy and enable it to capture more complex relationships between the variables.

To further improve our model's performance, we investigate the effect of removing redundant or noisy features from the dataset. We aim to identify the features that have a low contribution to the model's accuracy and remove them from the dataset. Additionally, we consider reconstructing the dataset in a balanced manner to ensure that the reconstructed dataset represents the majority and minority classes more accurately.

Also, we investigate whether combining the resulting models from our best previous solutions can lead to even better performance. We explore the possibilities of ensemble learning, where we combine different models to obtain a more accurate prediction.

Finally, we evaluate the strategies using various metrics to comprehensively understand their performance and determine the most suitable strategy for an imbalanced dataset. By analyzing different metrics, we aim to identify the strengths and limitations of each strategy and gain insights into what each metric reflects. This evaluation will enable us to make informed decisions and select the optimal strategy for dealing with imbalanced data.

Our study aimed to develop a reliable and accurate model to predict missing information about the type of metiers that can be linked to each landing and gain insights into the overall dataset.

\subsection{Methods}\label{subsec:methods}

This section describes the methods used to address the research questions, including pre-processing, predictive solutions, and evaluation and estimation methods.

\subsubsection{Pre-processing Solutions}\label{subsubsec:pre_pro_sol}

Pre-processing strategies are essential to improve model performance. Here we detail the ones used in this work. 

\begin{itemize}
    \item \textbf{Context-related variables:} To investigate whether adding context-related variables can improve the performance of machine learning models, the study added information on the weight of each fish group classification over the past six months. This information was relevant since the weight of fish is likely to vary over time, and including it could help improve the accuracy of the models. To calculate the new variables, we first identified the date of each observation and then computed the maximum, minimum, and mean weight of each class over the past six months. We then added these new variables to our dataset, which increased the number of features available for modeling.
    \item \textbf{Feature Selection:} Feature selection is a technique used to improve the performance of machine learning models by removing noisy or irrelevant features from the dataset. In this study, the Boruta algorithm was employed, an all-relevant feature selection wrapper algorithm that ranks the features according to their importance and determines which features are relevant, irrelevant, or uncertain. 
    \item \textbf{Balanced Reconstructions:} Resampling strategies modify the original data distribution to meet specific user-defined criteria. The study evaluated the performance of different sampling techniques, including undersampling of the majority classes, oversampling of the minority classes, and two hybrid techniques that combined both approaches. The functions and the corresponding parameters are described in~\ref{anexx2}. 
\end{itemize}

\subsubsection{Predictive Solutions}\label{subsubsec:predsol}

Supervised learning models are commonly used in classification solutions \cite{ps1}. Tree-based approaches are a popular class of algorithms that offer good model interpretability, making them suitable for our task \cite{ps2}. Therefore, for our study, we have chosen three tree-based approaches:

\begin{itemize}
    \item \textbf{Decision tree classifier}: can be used to create a hierarchical partition of the training data space, where the constraints on attribute values are used to split the data~\cite{ps3}. This iterative process continues until leaf nodes are formed, each containing a small number of records that can be used for classification purposes;
    \item \textbf{Random Forest}: The algorithm is based on an ensemble technique, where several trees are trained and then combined to make predictions~\cite{ps4};
    \item \textbf{Boosting}: This algorithm is a meta-heuristic approach based on machine learning, which aims to reduce bias and variance in supervised learning~\cite{ps5}. It belongs to a family of machine learning methods that transform a weak learner into a stronger one.
\end{itemize}

To ensure that our work is easily replicable, we used the implementations of these tools available in the free and open-source R environment. Concerning the parameter settings for each of these methods. We carried out a preliminary test to search for optimal parameterization (i.e., the setting that obtains the best possible results within a certain set of values of the parameters). The search for optimal parameters was carried out for each combination machine learning model - time window, described in the~\ref{subsubsec:eval_n_est}, and the results are detailed in Annex~\ref{anexx1}.

\subsubsection{Evaluation and Estimation}\label{subsubsec:eval_n_est}

To evaluate the performance of predictive models on this dataset, it was decided to consider different time windows for training and testing. So it was decided to use different time windows for training and testing. We selected three methods: sliding, growing, and full.

\begin{itemize}
    \item \textbf{Sliding Window:} Trains the model on two years of data and tests it on three months, with the two-year period being before the three-month test period. We increment the window by one month for training and testing over the years;
    \item \textbf{Growing Window:} Trains the model on all the available data before the test period (which is also three months), starting with a two-year window and incrementing it by one month for both training and testing. 
    \item \textbf{Full Window:} Both methods are commonly used in time series analysis, but as we needed to fill in missing values, we thought of using another approach called the full window method. This method considers all the data before and after the three-month test period, with increments of three months over the years for evaluation.
\end{itemize}

\subsection{Results}\label{subsec:results}

This section provides an overview of the answers to the research questions proposed in this study. We aim to comprehensively understand the effectiveness of the different models and techniques used in this study.

\subsubsection{\textbf{ML models Performance Across Time Windows}}
\begin{figure*}[htbp]
\centerline{\includegraphics[width=1.0\linewidth]{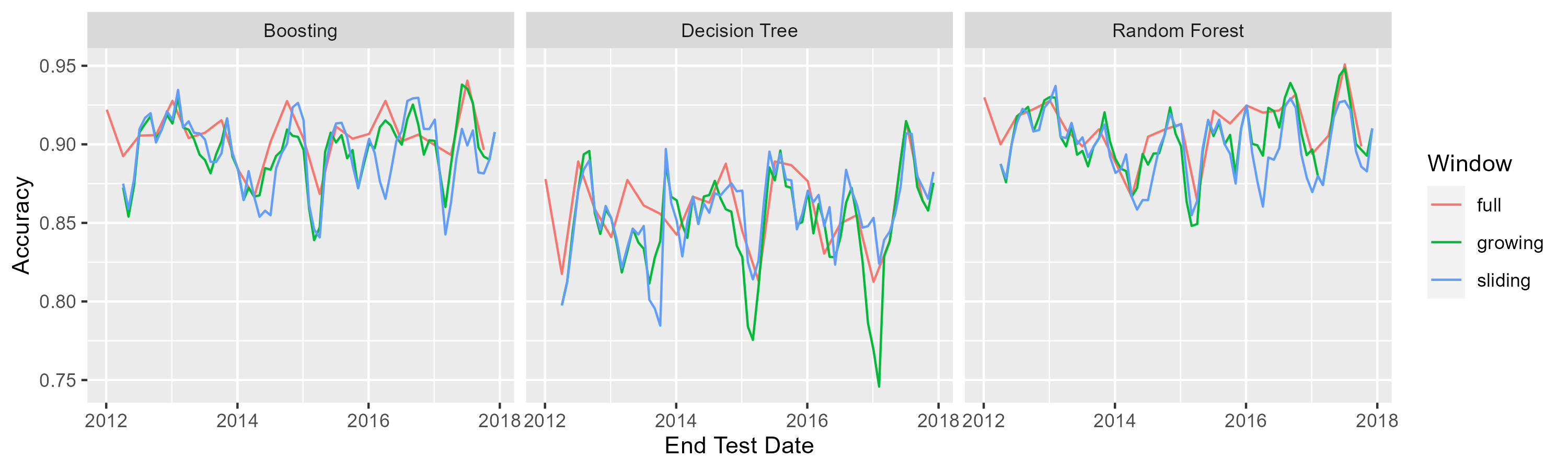}}
\caption{Machine learning methods and time windows}
\label{fig_methods_windows}
\end{figure*}

The first research question we aimed to address was to determine the optimal combination of predictive solution and time window for training. To this, we analyzed the accuracy of each combination of predictive solutions and time window methods over the years, as illustrated in figure \ref{fig_methods_windows}. The results indicate that the full-time window produced more stable and better results across all models. Although both the boosting and random forest models exhibited good performance, we used the boosting model with the full-time window method. We made this decision based on its consistent performance and stability over time and its ability to handle complex relationships between variables. We now call this the Base strategy.  

\subsubsection{\textbf{Add New Features}}
We aimed to investigate whether adding new features to the model could improve its performance. To test this, we created a new data set by introducing features explained in~\ref{subsubsec:pre_pro_sol} and applied the base strategy. However, the results showed that the model with the new features yielded the same mean accuracy as the model without them (90\%). Nevertheless, we believe that including these features would significantly impact the model's performance on a larger version of this data set. Therefore, we decided to use the boosting algorithm with the full window evaluation and the new features for our subsequent analyses. In Table~\ref{tab_results}, we present the results of the strategies tested in this section in the new data set. 

\begin{table*}[htbp]
\caption{Accuracy of each metier class per method on the data set with new features}
\begin{center}
\scalebox{0.8}{
\begin{tabular}{|c|c|c|c|c|c|}
\hline
\textbf{Metiers}&\multicolumn{5}{|c|}{\textbf{Methods}} \\
\cline{2-6} 
& \textbf{\textit{Base}} & \textbf{\textit{ImpSamp}} & \textbf{\textit{OverSamp}} &   \textbf{\textit{Ensemble}} &  \textbf{\textit{Feature Selection} }\\
\hline
 FPO-CRU & 0\% & 0\% & 0\% & 0\%  & 0\%\\
 FPO-PB & 19,5\% & 50\% & 19,5\% & \textbf{52,1\%} & 15,2\% \\
 GNS-PB & 61,9\% & 56,7\% & 61,3\% & \textbf{62,4\%}& 61,6\% \\
 LHP-CEF & \textbf{99,3\%} & 99,1\% & \textbf{99,3\%} & 99,2\% & \textbf{99,3\%} \\
 LHP-PB & \textbf{91,8\%} & 85\% & 91,3\% & 88,2\%  & \textbf{91,8\%} \\
LHP-PBC & 0\% & \textbf{27\%} & 9\% & 17,9\%  & 0\% \\
LHP-TUN & \textbf{99,4\%} & 99\% & \textbf{99,4\%} & \textbf{99,4\%} & \textbf{99,4\%} \\
LLD-GPP & 64,5\% & 48,3\% & 64,5\% & \textbf{66.7\%} & 61,5\% \\
LLD-PP & 97,2\% & 92,6\% & \textbf{98,1\%} & 94,5\% & 97,2\% \\
LLS-DEEP & 22,9\% & 78,7\% & 27,2\% & \textbf{80.5\%}  & 22\% \\
LLS-PD & \textbf{86,6\%} & 72,4\% & 86\% & 71,8\% & 86,4\% \\
PS-PB & 0\% & \textbf{62,7\%} & 6,9\% & 60,4\% & 0\% \\
PS-PPP & 96,6\% & 96,3\% & 96,6\% & \textbf{96,7\%} & 96,6\% \\
\hline
Final Accuracy  & \textbf{91\%} & 87\% & 90,8\% & 88,2\% & 88,7\% \\
\hline
Geometric Mean &  \textbf{71.7\%} & 70.1\% & 49.1\% & 70.6\% & 69.8\% \\
\hline
Balanced Mean & \textbf{91.02} & 86.97 & 90.84 & 88.21 & 90.93 \\
\hline
Imbalanced Mean & 54.06 & 65.07 & 55.69 & \textbf{66.80} & 53.34 \\
\hline
\multicolumn{6}{l}{}
\end{tabular}
}
\label{tab_results}
\end{center}
\end{table*}

\subsubsection{\textbf{Feature selection}}
The new data set has 180 variables, but we suspected some might be affecting the model's performance. To investigate this, we performed feature selection as described in the \ref{subsubsec:pre_pro_sol} and ultimately identified 98 relevant variables, 76 irrelevant ones, and two uncertain variables. We used only these 98 selected variables for our subsequent analysis.

In \ref{tab_results}, it is noteworthy that the feature selection didn't improve any of the classes relative to the model1, and the final accuracy even decreased. This suggests that even the least relevant variables play an important role in distinguishing between classes.

\subsubsection{\textbf{Resampling}}
Table \ref{tab_results} reveals that certain minority classes, such as FPO-CRU, FPO-PB, LLS-DEEP, and PS-PB, presented in table \ref{tab_metiers}, had low accuracy, with some even having an accuracy of zero. On the other hand, the majority classes, including LLS-PD, LHP-CEF, and LHP-PB, demonstrated good performance, with an accuracy greater than 86\%. In response to this imbalance, we constructed a more balanced dataset, focusing on these classes, except for FPO-CRU, which occurred only twice in the entire dataset.

We employed various strategies to select the best combination of parameters for undersampling and oversampling, as detailed in Appendix \ref{anexx2}. This resulted in two optimal models, OverSamp and ImpSamp, and their performances for each class are illustrated in Table \ref{tab_results}.

The ImpSamp model was highly effective in improving the performance of the four proposed minority classes, as shown in Table \ref{tab_results}. On the other hand, OverSamp showed only a slight improvement in LHP-TUN and LLD-PP, which were not our primary focus. These results suggest that balancing the data set positively impacts the performance of minority classes, albeit with a trade-off of slightly reduced accuracy in the majority classes.

\subsubsection{\textbf{Ensemble}}
After analyzing the results obtained from the ImpSamp and Model1 models, we considered creating a compensatory ensemble model that could maintain the accuracy of the majority classes while preserving the gains made in the minority classes. To explore this option, we designed a new ensemble model named the Ensemble model.

We made predictions using the Model1 and ImpSamp strategies in the Ensemble model. We retained the results from the ImpSamp approach if it predicted the FPO-PB, LLS-DEEP, or PS-PB metiers. However, if the ImpSamp approach did not predict any of these metiers, we kept the results from the Model1 approach.

We evaluated the performance of the Heuristic model and compared it to the results obtained by the previous models. As shown in Table \ref{tab_results}, the Ensemble model maintained the accuracy gained with the ImpSamp approach without compromising the performance of the majority classes with the Base approach.  This method presents accuracy higher than 52\% for all, except for FPO-CRU and LHP-PBC, which only occurred in 2017, with very low occurrences in data.

\subsubsection{\textbf{Metric Evaluation}}
In addition to evaluating the accuracy of each classification strategy individually, we also compared their results using three different metrics: the geometric mean of the accuracy of each class, the average of each class weighted by its proportion in the dataset (balanced mean), and the average of each class weighted by the inversion of its proportion in the dataset (imbalanced mean). Table~\ref{tab_results} showed that the Base strategy outperformed all metrics except for the imbalanced mean. This metric gives more weight to underrepresented classes, which had the best performance from the Ensemble strategy. As mentioned, the Ensemble strategy is particularly effective in predicting minority classes. Therefore, it can be concluded that the Ensemble strategy can provide the most accurate predictions for imbalanced datasets, while the Base strategy is more suitable for balanced datasets.

\section{Discussion}\label{sec:discussion}  

It seems that the models' performance is very high for the most selective gears, those gears which are very specific to catch a small number of species, e.g., the LHP-CEF for catching squids, the LHP-TUN for catching tunas or the PS-PPP to catch small pelagics like the \textit{Trachurus picturatus}.

Table \ref{tab_iconic_species} provides insights into the three iconic species associated with each class of fishery art, along with their respective percentages of total weight within that class. On the other hand, Table \ref{confusion_matrix} displays the confusion matrix resulting from the ensemble method. By examining these tables, we can identify similarities between fishery landings and explore the misclassification patterns.

\begin{table*}
\caption{Three Iconic Species of each metier}
\begin{center}
\resizebox{18cm}{!}{%
\begin{tabular}{|c|c|c|c|} 
 \hline
 \textbf{Metier} & \textbf{First} & \textbf{Second} & \textbf{Third} \\ 
 \hline
FPO-CRU & DEMERSAIS (93.94\%) & CRUSTACEOS (6.06\%) & \\
 \hline
FPO-PB & DEMERSAIS COSTEIROS (52.64\%) & DEMERSAIS (34.81\%) & MOLUSCOS (5.76\%) \\
 \hline
GNS-PB & DEMERSAIS COSTEIROS (74.31\%) & PELAGICOS COSTEIROS (20.31\%) & DEMERSAIS TALUDE (2.09\%) \\
 \hline
LHP-CEF & MOLUSCOS (99.67\%) & BENTICO-PELAGICOS TALUDE (0.11\%) & DEMERSAIS TALUDE (0.09\%) \\
 \hline
LHP-PB & BENTICO-PELAGICOS TALUDE (36.92\%) & DEMERSAIS TALUDE (23.03\%) & DEMERSAIS COSTEIROS (16.48\%)  \\
 \hline
LHP-PBC & PELAGICOS COSTEIROS (96.85\%) & DEMERSAIS COSTEIROS (2.65\%) & MOLUSCOS (0.50\%) \\
 \hline
LHP-TUN & TUNIDEOS (99.992\%) & PELAGICOS COSTEIROS (0.003\%) & GRANDES PELAGICOS MIGRADORES (0.003\%) \\
 \hline
LLD-GPP & GRANDES PELAGICOS MIGRADORES (96.52\%) & DEMERSAIS (0.41\%) & BENTICO-PELAGICOS TALUDE (0.38\%) \\
 \hline
LLD-PP &  ESPECIES PROFUNDIDADE (99.879\%) & DEMERSAIS TALUDE (0.119\%) & OUTRAS SPP (0.002\%) \\
 \hline
LLS-DEEP &  ESPECIES PROFUNDIDADE (51.16\%) & DEMERSAIS TALUDE (30.76\%) & BENTICO-PELAGICOS TALUDE (16.71\%) \\
 \hline
LLS-PD & DEMERSAIS TALUDE (46.79\%) & BENTICO-PELAGICOS TALUDE (25.75\%) & DEMERSAIS (15.52\%) \\
 \hline
PS-PB &  PELAGICOS COSTEIROS (35.82\%) & PEQUENOS PEIXES DEMERSAIS COSTEIROS (35.79\%) & DEMERSAIS COSTEIROS (28.40\%) \\
 \hline
PS-PPP &  PEQUENOS PELAGICOS (99.79\%) & MOLUSCOS (0.11\%) & TUNIDEOS (0.04\%) \\
 \hline
\end{tabular}%
}
\end{center}
\label{tab_iconic_species}
\end{table*}

\begin{table*}[htbp]
\caption{Confusion Matrix generate with ensemble strategy}
\begin{center}
\scalebox{0.8}{
\begin{tabular}{|c|c|c|c|c|c|c|c|c|c|c|c|c|c|}
\hline
\textbf{True}&\multicolumn{13}{|c|}{\textbf{Predicted}} \\
\cline{2-14} 
& FPO-CRU & FPO-PB & GNS-PB & LHP-CEF & LHP-PB & LHP-PBC &  LHP-TUN & LLD-GPP & LLD-PP&  LLS-DEEP & LLS-PD & PS-PB & PS-PPP\\
\hline
FPO-CRU & \cellcolor{blue!25}0 & 0 & 0 & 0 & 2 &  0 & 0 & 0 & 0 & 0 & 0 & 0 & 0 \\
FPO-PB & 0 & \cellcolor{blue!25}24 & 0 & 1 & 16 & 0 & 0 & 0 & 0 & 0 & 3 & 2 & 0 \\
GNS-PB & 0 & 2 & \cellcolor{blue!25}231 & 0 & 132 & 0 & 0 & 0 & 0 & 0 & 1 & 4 & 0 \\
LHP-CEF & 0 & 1 & 0 & \cellcolor{blue!25}3377 & 20 & 0 & 0 & 0 & 0 & 0 & 3 & 0 & 0 \\
LHP-PB & 0 & 57 & 14 & 26 & \cellcolor{blue!25}4100 & 18 & 3 & 1 & 0 & 74 & 288 & 64 & 4 \\
LHP-PBC & 0 & 0 & 0 & 2 & 21 & \cellcolor{blue!25}5 & 0 & 0 & 0 & 0 & 0 & 0 & 0 \\
LHP-TUN & 0 & 0 & 0 & 1 & 5 & 0 & \cellcolor{blue!25}1020 & 0 & 0 & 0 & 0 & 0 & 0 \\
LLD-GPP & 0 & 0 & 0 & 0 & 4 & 0 & 4 & \cellcolor{blue!25}24 & 0 & 0 & 4 & 0 & 0 \\
LLD-PP & 0 & 0 & 0 & 0 & 0 & 0 & 0 & 0 & \cellcolor{blue!25}103 & 6 & 0 & 0 & 0 \\
LLS-DEEP & 0 & 0 & 0 & 0 & 9 & 1 & 1 & 0 & 1 & \cellcolor{blue!25}186 & 33 & 0 & 0 \\
LLS-PD & 0 & 8 & 0 & 3 & 331 & 0 & 5 & 1 & 2 & 438 & \cellcolor{blue!25}2004 & 0 & 0 \\
PS-PB & 0 & 0 & 2 & 0 & 15 & 0 & 0 & 0 & 0 & 0 & 0 & \cellcolor{blue!25}26 & 0 \\
PS-PPP & 0 & 2 & 0 & 22 & 15 & 0 & 1& 0 & 0 & 0 & 0 & 12 & \cellcolor{blue!25}1515 \\
\hline
\multicolumn{14}{l}{}
\end{tabular}
}
\label{confusion_matrix}
\end{center}
\end{table*}

For example, an analysis of the data reveals that GNS-PB is predicted as LHP-PB in 35\% of the cases. This can be attributed to the low occurrence of the GNS-PB fishery art in the dataset (as shown in Table~\ref{tab_metiers}), as well as the presence of common fishery species such as Demersais Costeiros and Demersais Talude. Similarly, LLS-PD is observed to be distributed between LHP-PB and LLS-DEEP. This can be explained by the fact that these three arts share iconic fishery species like Demersais Talude and Bentico Peleagicos, leading to similarities in their predicted classifications.

Understanding the similarity between fishery arts is crucial for assessing the impact of fishing activities on fish populations and their habitats. Researchers can identify common trends and drivers of changes in fish stocks by analyzing catch composition, fishing techniques, and spatial distribution across different arts. This knowledge is essential for evaluating the sustainability of fishing practices and implementing measures to prevent overfishing and protect vulnerable species.

As part of future work, we aim to enhance the predictiveness of our model. One approach we are considering is to use a graph-based representation of the total weight of each fish species over time, based on the work \cite{c1}. This would allow us to extract patterns that can be used as features to improve the model's accuracy further. 

\section{Conclusions}\label{sec:conclusion}

In conclusion, our research addresses the crucial problem of incomplete datasets and missing data in fisheries by employing machine learning techniques to fill in missing information in the LOTAÇOR/OKEANOS-UAc fishery landings dataset. The study demonstrates the effectiveness of this approach in enhancing dataset completeness and uncovering valuable insights into fishery trends. The significance of our research lies in its implications for actionable policy-making and real-time sustainability decisions, as it enables decision-makers to make informed choices for resource management and conservation measures based on a more comprehensive and reliable dataset. By leveraging the potential of machine learning, we can contribute to the sustainable management of fisheries and the preservation of marine ecosystems by formulating evidence-based policies and making informed decisions in real time.


\section*{Acknowledgments}

This work received national funds through the FCT – Foundation for Science and Technology, I.P., under the project UIDB/05634/2020 and UIDP/05634/2020 and through the Regional Government of the Azores through the project M1.1.A/FUNC.UI\&D/003/2021-2024. We also acknowledge the LOTAÇOR, S.A., a regional public company responsible for the fish auctions and fisheries landings statistics, and João Santos (OKEANOS - UAc), which manages the LOTAÇOR/OKEANOS-UAC fishing landings database.

\section{Code Availability}
The code used in this work is available: 
\url{https://github.com/brendacnogueira/datareconstruction.git}

\appendix
\section*{Annex 1}\label{anexx1}

The following list describes the parameters tested in each machine learning algorithm in Section~\ref{subsubsec:eval_n_est}.
\begin{itemize}
    \item \textbf{Decision Tree}: maximum depth: 10
    \item \textbf{Random Forest}: number of trees:\{250, 500, 750\}
    \item \textbf{Boosting}: 
        \begin{itemize}
        \item maximum depth:\{2,10,20\}
        \item number of rounds:\{50,75,100\}
        \item early stopping rounds: 20 
        \end{itemize}
\end{itemize}

The table \ref{tab_anexx1} presents the best parameters for Random Forest and Boosting in each time window. We calculated the maximum and minimum accuracy for each parameter over time to select the optimal parameters and then computed the difference between them. We ranked the parameters based on this difference to determine the best-performing parameter for each method and time window combination.
\begin{table}
\caption{Best Parameters For Random Forest and Boosting for each window strategy}
\begin{center}
\begin{NiceTabular}{lll}[hvlines]
  Method & Window & Parameters \\ 
  \Block{3-1}{Random Forest} & Sliding & num.trees=750 \\ 
  & Growing & num.trees=250  \\  & Full & num.trees=250 \\ 
  \Block{3-1}{Boosting} & Sliding & max depth= 10, nrounds= 100\\ 
  & Growing & max depth= 10, nrounds= 75  \\  & Full & max depth= 10, nrounds= 50
\end{NiceTabular}
\end{center}
\label{tab_anexx1}
\end{table}

\section*{Annex 2}\label{anexx2}

In this study, we tested various functions for resampling techniques. We employed two methods to determine the optimal percentage of undersampling and oversampling for each class.

The first method involved setting the same percentage of undersampling and oversampling for all classes. The parameters tested are:

\begin{itemize}
    \item \textbf{Undersampling Percentage}: \{0.1 to 0.9, with increments of 0.1\}
    \item \textbf{Ooversampling Percentage}:\{0.25,0.5,0.75,1,2,3,4\}
\end{itemize}

The second method tested various combinations of oversampling and undersampling values for each class. We randomly selected combinations of the aforementioned values for testing.

To find the optimal combination of parameters and functions, we utilized two evaluation metrics across the Boosting algorithm and the full window evaluation: accuracy and a metric called TP, which is the sum of the individual accuracy of each class. The best two resampling strategies are in the \ref{tab_resampling}. The first strategy was denominated as Impsamp and the second as OveSamp.

\begin{table} 
\caption{Best resampling strategies}
\begin{center}
\resizebox{\columnwidth}{!}{
  \begin{tabular}{|c|c|c|c|c|c|c|c|}
    \hline
    \multirow{2}{*}{Function}  &
      \multicolumn{3}{c|}{Undersampling (\%)} &
      \multicolumn{4}{c|}{Oversampling (\%)} \\
    &  LLS-PD & LHP-CEF & LHP-PB & FPO-PB & LHP-PBC & LLS-DEEP & PS-PB\\
    \hline
     WERCSClassif & 0.3 & 0.3 & 0.3 & 4 & 4 & 4 & 4\\
    \hline
     RandOverClassif &  &  &  & 1.25 & 4 & 1.5 & 1.75\\
     \hline
  \end{tabular}
  }
  \label{tab_resampling}
  \end{center}
\end{table}

\end{document}